# Uncertainty Prediction Neural Network (UpNet): Embedding Artificial Neural Network in Bayesian Inversion Framework to Quantify the Uncertainty of Remote Sensing Retrieval

*Dasheng Fan[1,2], Xihan Mu[1,2,] [*] Yongkang Lai[1,2], Donghui Xie[1,2] and Guangjian Yan[1,2]*

[1]State Key Laboratory of Remote Sensing Science, Faculty of Geographical Science, Beijing Normal University, Beijing 100875, China

[2]Beijing Engineering Research Center for Global Land Remote Sensing Products, Faculty of Geographical Science, Beijing Normal University, Beijing 100875, China

* Corresponding author: muxihan@bnu.edu.cn

**Abstract:** For the retrieval of large-scale vegetation biophysical parameters, the inversion of radiative transfer models (RTMs) is the most commonly used approach. In recent years, Artificial Neural Network (ANN)-based methods have become the mainstream for inverting RTMs due to their high accuracy and computational efficiency. It has been widely used in the retrieval of biophysical variables (BV). However, due to the lack of the Bayesian inversion theory interpretation, it faces challenges in quantifying the retrieval uncertainty, a crucial metric for product quality validation and downstream applications such as data assimilation or ecosystem carbon cycling modeling. This study proved that the ANN trained with squared loss outputs the posterior mean, providing a rigorous foundation for its uncertainty quantification, regularization, and incorporation of prior information. A Bayesian theoretical framework was subsequently proposed for ANN-based methods. Using this framework, we derived a new algorithm called Uncertainty Prediction Neural Network (UpNet), which enables the simultaneous training of two ANNs to retrieve BV and provide retrieval uncertainty. To validate our method, we compared UpNet with the standard Bayesian inference method, *i.e.*, Markov Chain Monte Carlo (MCMC), in the inversion of a widely used RTM called ProSAIL for retrieving BVs and estimating uncertainty. The results demonstrated that the BVs retrieved and the uncertainties estimated by UpNet were highly consistent with those from MCMC, achieving over a million-fold acceleration. These results indicated that UpNet has significant potential for fast retrieval and uncertainty quantification of BVs or other parameters with medium and high-resolution remote sensing data. Our Python implementation is available at: https://github.com/Dash-RSer/UpNet.

**Keywords:** Artificial Neural Networks, Inverse Problem, Remote Sensing Retrieval, Radiative Transfer Models, Biophysical Parameters, Uncertainty Quantification

## I. Introduction

With the intensification of global climate change and human activities, monitoring surface vegetation biophysical variables (BVs) on a large scale has become increasingly important, as they can be used to assess crop growth and the condition of ecological environments, addressing food security and global climate change issues (Baret et al., 2008; Weiss et al., 2020; Zemp et al., 2022). Remote sensing retrieval, with its advantages of covering large areas at low cost, has been widely used for inferring surface BVs from observations acquired by satellite sensors (Fang et al., 2019).

Remote sensing retrieval is typically categorized into two types: empirical methods and physical methods (Wang et al., 2022). Empirical methods rely on field measurement data to calibrate statistical models, establishing an empirical mapping from sensor observations to BVs. However, this approach is limited by the spatial range of data collection and is not suitable for large-scale BV retrieval. In contrast, physical methods utilize physical principles to construct models that relate BVs to observations, which is called radiative transfer model (RTM). After obtaining observations, BVs can be retrieved by solving the inverse problems of RTM. Due to the universality of physical laws, RTM inversion is well-suited for large-scale BV retrieval and become a key area of remote sensing retrieval research. Typically, there are three main methods for RTM inversion: iterative optimization methods (Bacour et al., 2002; Combal et al., 2003; Fang et al., 2003), lookup table methods (LUT) (Knyazikhin et al., 1998; Weiss et al., 2000), and the Artificial Neural Network (ANN)-based method (Danner et al., 2021). They have been used for median-resolution BV retrieval, such as the famous Moderate Resolution Imaging Spectroradiometer (MODIS) leaf area index (LAI) product (Knyazikhin et al., 1998) and GEO LAI product (Baret et al., 2007). Recently, there has been an increasing demand for BV retrieval using high-resolution imagery (Zérah et al., 2024). However, both iterative optimization and LUT require processing each pixel individually (Verrelst et al., 2019), which makes it challenging to meet the speed requirements for high-resolution BVs retrieval (Zérah et al., 2024). The ANN-based method uses training datasets simulated by RTMs to train the ANN model, which is then applied to actual satellite observations (Verrelst et al., 2015). This method combines the physical knowledge of RTMs with the extremely high computational efficiency of ANN, often processing an entire remote sensing image almost instantly (Verrelst et al., 2019). Therefore, this method has gradually become the mainstream approach for RTM retrieval.

In recent years, retrieval uncertainty, as a critical indicator of retrieval quality, has gained increasing attention in addition to retrieving BVs (Fang et al., 2019). It can be used to assess the quality of satellite products (Pinty et al., 2011; Fang et al., 2012; Fang et al., 2021; Yan et al., 2021; Wang et al., 2023), thereby facilitating the improvement of satellite products. Moreover, retrieval uncertainty serves as a crucial input for downstream products such as data assimilation (Chen et al., 2020), ecosystem carbon cycling model (Viskari et al., 2015), and the construction of ecohydrologic error models (Tang et al., 2019). To obtain the uncertainty of the retrieval, the standard practice is to use standard deviation of the posterior distribution as the retrieval uncertainty from Bayesian inversion theory[1] (Wang et al., 2022), which is also a requirement of GCOS (Zemp et al., 2022). However, due to the lack of an analytical solution for the posterior distribution, it is very difficult to obtain its standard deviation. To now, researchers can only empirically select a certain percentage of results in LUT to statistically calculate the standard deviation as the uncertainty (Combal et al., 2002; Koetz et al., 2005; Richter et al., 2012), or linearize the model to obtain an approximate uncertainty in iterative optimization (Rogers, 2000). The uncertainty obtained from these methods is somewhat arbitrarily defined (Disney et al., 2018), and their rationality in downstream applications is questionable.

For ANN-based method, although it has shown great potential in BV retrieval, there is still no reasonable method to obtain the posterior uncertainty due to the lack of Bayesian theoretical interpretation. This not only affects the quantification of retrieval uncertainty but also poses difficulties in utilizing prior knowledge and regularizing the retrieval. Recently, some attempts have been made to enable ANNs to output uncertainty, *e.g.*, MC (Monte Carlo) dropout (Martínez-Ferrer et al., 2022), and variational auto encoder (VAE) (Svendsen et al., 2024). However, these methods often involve strong assumptions and require lots of model runs for a single prediction, and are severely limited in remote sensing applications. Gaussian process regression (GPR) has been explored as an alternative to ANNs due to its ability to output uncertainty (Verrelst et al., 2012; Verrelst et al., 2013; Verrelst et al., 2015; Estévez et al., 2021; Estévez et al., 2022). However, the computational complexity of GPR increases rapidly with simulated datasets, limiting its application on RTM inversion (Camps-Valls et al., 2016).

---

[1]In this paper, we consider retrieval uncertainty to be equivalent to the posterior standard deviation, without distinguishing between these two concepts. Hereafter, we will refer to both as uncertainty.

Another emerging class of RTM inversion methods is known as Bayesian parameter inference (Shiklomanov et al., 2016; Varvia et al., 2017; Wang et al., 2022), which obtains the statistics of the posterior distribution through Bayesian theory to perform retrieval and uncertainty quantification. The representative algorithm is the well-known Markov Chain Monte Carlo (MCMC) method. MCMC constructs a Markov chain whose stationary distribution is the posterior distribution, and subsequently samples the posterior distribution of parameters to obtain posterior samples (Hastings, 1970). By conducting statistical analysis on these posterior samples, retrieval results and posterior uncertainty can be output. MCMC is grounded in solid theoretical foundations and is a standard method for uncertainty quantification in statistics and machine learning (Bishop, 2006). However, although the remote sensing community has made initial attempts to use MCMC for BV retrieval (Wang et al., 2022), this method requires a large number of forward model runs for each pixel, leading to a very high computational cost and making it unsuitable for producing large-scale BV products (Baret et al., 2008). To summarize, up to now, there has been no method capable of rapidly and accurately quantifying retrieval uncertainty.

In this paper, we aim to combine the ANN-based retrieval with the Bayesian parameter inference to develop a retrieval method with high accuracy, high computational efficiency, and effective uncertainty quantification. Our main contributions are as follows:

(I) to proof ANN-based method can predict the posterior mean of BV, and subsequently establish a Bayesian theoretical framework for ANN-based method.

(II) to propose a simple, robust, and efficient algorithm called Uncertainty predict Neural Network (UpNet) to retrieve BVs and provide retrieval uncertainty simultaneously.

(III) to compare the retrieved BV and retrieval uncertainty obtained from the UpNet with those from MCMC based on simulated and real-world datasets.

## II. Method

### 2.1 Background knowledge of RTM inversion

#### 2.1.1 Bayesian inversion theory

For notations, we use $\theta \in \mathbb{R}^m$ for the m-dimensional BVs in the RTM, $\theta^k \in \mathbb{R}$ for the k-th dimension of $\theta$ which is the BV we want to retrieve, $r \in \mathbb{R}^n$ for the n-dimensional observation, the probability density of a random variable is denoted as $p(\cdot)$, and RTM is denoted as $f(\theta)$. For example, if there are two parameters of RTM $f(\theta)$: LAI and average leaf angle (ALA). Then, $\theta$ is the vector $[LAI, ALA]^T$, and the $\theta^1$ is $LAI$ and $\theta^2$ is ALA. Although our method is not limited to optical remote sensing, in this paper, we use optical remote sensing of vegetation as an example, where the observations $r$ is the canopy reflectance and $f(\theta)$ denotes the canopy RTM.

Our method is based on the Bayesian inversion framework, which is the general inversion theory framework and naturally integrates observations, models, and prior information. It can encompass most inversion methods of RTMs, such as the iterative optimization method and LUT method. In the Bayesian framework, the observations and parameters are both viewed as a random vector, then the retrieval process is expressed by Bayes' formula (Bishop et al., 2006):

$$p(\theta|r) = \frac{p(r|\theta)p(\theta)}{p(r)}. \qquad (1)$$

Here, $p(\theta)$ is the prior distribution of BVs, $p(r|\theta)$ is likelihood function, $p(\theta|r)$ is posterior distribution of BVs, $p(r)$ is the normalization constant, which makes sure that the right-hand side of the equation is a probability density. The prior distribution is the prior knowledge of the BVs, for example, maximum and minimum values of the BVs in the field measurement or the physical bounds of the BVs, which could be viewed as a uniform prior distribution. If the mean and variance are known, the Gaussian prior could be used. In this study, we assume that the prior is uniform or (truncated) Gaussian, which is commonly used in previous researches (Baret et al., 2007; Jiang et al., 2022). In some studies, the prior distribution can be set in more complex forms, such as removing the assumption of independence between different BV distributions and considering BV correlations (Wang et al., 2022). In this study, we adopt the most common independent prior distribution to seek results under the most general settings. The likelihood function is the conditional distribution of observations, which is a function of the BVs (Rogers, 2000; Kaipio et al., 2006):

$$p(r|\theta) = f(\theta) + \epsilon = \mathcal{N}(f(\theta), \Sigma_r) = \frac{1}{(2\pi)^{\frac{n}{2}}|\Sigma_r|^{\frac{1}{2}}} \exp\left\{-\frac{1}{2}(r - f(\theta))^T \Sigma_r^{-1}(r - f(\theta))\right\}. \qquad (2)$$

Here the $f(\theta)$ is the RTM (*e.g.*, ProSAIL model), the $\epsilon \sim N(0, \Sigma_r)$ is the noise, $\Sigma_r$ is the covariance matrix of the observation noise. Finally, the posterior distribution $p(\theta|r)$ expresses the distribution of BVs after retrieval with observations.

In the complete Bayesian framework, the results of BV estimation are represented by the posterior distribution (Bishop, 2006). In remote sensing applications, a specific value of BV retrieval is typically required; therefore, it is necessary to summarize the posterior distribution into a specific value, known as point estimation (Lehmann and Casella, 1998). There are typically three ways to summarize the posterior distribution of BV into a point estimation: maximum a posteriori (MAP) estimation, posterior mean estimation, and posterior median estimation (Wasserman, 2013). They use the maximum probability value, mean value, and median value of the posterior distribution as the estimates, respectively. However, since the forward model $f(\theta)$ is typically nonlinear (*e.g.*, ProSAIL), an analytical expression for the posterior distribution cannot be obtained, making it difficult to estimate the posterior mean and median. Therefore, MAP is the most widely used approach since it could be formulated as an optimization problem and could be solved by optimization algorithms. The LUT method and iterative optimization method both belong to the MAP estimation. However, ANN-based methods have not been provided with a corresponding Bayesian interpretation.

One major advantage of Bayesian inversion theory is that it provides a method for quantifying uncertainty. As previously mentioned, the estimated uncertainty is defined as the standard deviation of the posterior distribution, denoted as

$$SD = \sqrt{\int_\theta p(\theta|r)\left(\theta - \mathbb{E}[\theta^k|r]\right)^2 d\theta} = \sqrt{\mathbb{V}[\theta^k|r]}. \qquad (3)$$

Here, $\mathbb{E}[\cdot]$ is the expectation of the random variable and $\mathbb{V}[\cdot]$ is the variance of the random variable.

**2.1.2 ANN-based method of RTM Inversion**

In this subsection, we introduce the basic workflow of ANN-based method of RTM inversion. This method first utilizes RTM to generate a training dataset to train an ANN. Specifically, to train the ANN, a distribution $p(\theta)$ is specified to generate samples of BVs, known as the training distribution. After obtaining $N$ training samples $\theta_i$, the RTM $f(\theta_i)$ is run to simulate the reflectance received by the sensor, and a noise sampled from $\epsilon$ in Eq. (2) is added to simulate the actual reflectance $r_i$. The simulation dataset is formed as $\mathcal{D} = \{(\theta_i, r_i)\}_{i=1}^{N}$. Then, with the ANN model $g(r)$, the objective function $J$ is minimized:

$$J = \frac{1}{N}\sum_{i=1}^{N} l\left(\theta_i^k, g(r_i)\right). \tag{4}$$

For the RTM inversion, the loss function $l$ has many choices. In this paper, the most commonly used squared loss $l_s$ is used:

$$l_s(\theta^k, g(r)) = \left(\theta^k - g(r)\right)^2. \tag{5}$$

With this squared loss function, the ANN $g(r)$ could be trained to retrieve BVs given the reflectance $r$. The ANN models typically consist of an input layer, multiple hidden layers, and an output layer. This paper adopts the widely used multilayer perceptron (MLP) architecture, where these layers are connected by a linear transformation and a non-linear activation function (Goodfellow et al.,2016). As a universal approximator, ANN can approximate any continuous function to any desired degree of accuracy (Hornik et al., 1989).

## 2.2 Theory foundation of UpNet algorithm

### 2.2.1 Overview of the theoretical framework

In this section, we developed an algorithm called UpNet, which simultaneously (I) retrieves BV and (II) provides the retrieval uncertainty. The UpNet consists of two ANNs, i.e., $g(r)$ and $u(r)$, to achieve (I) and (II), respectively. Goal (I) can be easily achieved by training an ANN ($g(r)$) using the loss function (Eq. (5)).

The realization of goal (II) depends on the other ANN $u(r)$ trained with loss function

$$l_u(\theta^k, u(r)) = \left(\left(\theta^k - g(r)\right)^2 - u(r)\right)^2. \tag{6}$$

Here, $g(r)$ is the first ANN that has been trained with loss function Eq. (5) for goal (I). In Section 2.2.2, we first proved that $g(r)$ actually predicts the posterior mean:

$$g(r) = \mathbb{E}[\theta|r]. \tag{7}$$

In Section 2.2.3, we proved that the second ANN $u(r)$ predicts the posterior variance $\mathbb{V}[\theta^k|r]$. Using the first ANN $g(r)$ and the second ANN $u(r)$, we could retrieve BV and providing retrieval uncertainty simultaneously. The overall workflow of UpNet is shown in Fig. 1.

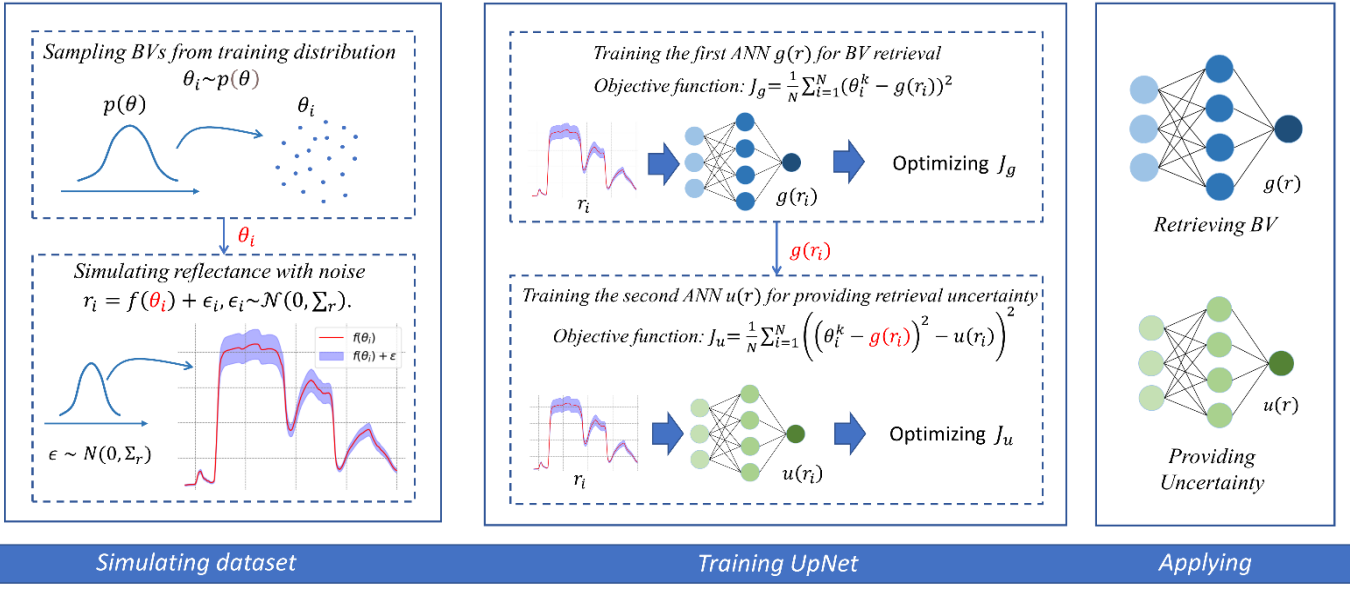

**Fig. 1.** The workflow of UpNet.

**2.2.2 ANN trained with squared loss approximately predicts the mean of posterior distribution**

In this subsection, we aim to prove that the first ANN $g(r)$ trained using loss function Eq. (5) predicts the posterior mean, *i.e.*, $g(r) = \mathbb{E}[\theta|r]$ (Eq. (7)). The key to the proof is establishing a Bayesian theoretical framework for ANN-based method. To achieve this, we introduce statistical decision theory, which is used to evaluate the properties of estimators in statistics. In statistical decision theory, the Bayes risk is commonly employed to evaluate the estimation error of an estimator. In remote sensing, the estimator is the retrieval algorithm. The Bayes risk $R(\delta)$ is defined as (Wasserman, 2004):

$$R(\delta) = \int_\theta p(\theta) \int_r p(r|\theta) \, l\left(h(\theta^k), \delta(r)\right) dr d\theta \tag{8}$$

Here, $\delta$ is the retrieval algorithm (estimator), $r$ is the reflectance, and $l$ is the loss function that measures the distance between the retrieved value $\delta(r)$ and the function $h(\theta^k)$. $h(\theta^k)$ could be any function related to our interested BV ($\theta^k$). If $l$ is the squared loss, then by Lehmann and Casella (1998) (see Lemma 1 in Appendix A.1), we have:

$$\mathbb{E}[h(\theta^k)|r] = \arg\min_\delta \int_\theta p(\theta) \int_r p(r|\theta) \left(\delta(r) - h(\theta^k)\right)^2 dr d\theta. \tag{9}$$

Let $h(\theta) = \theta$, which means that we are interested in the BV $\theta^k$ itself, we have:

$$\mathbb{E}[\theta^k|r] = \arg\min_\delta \int_\theta p(\theta) \int_r p(r|\theta)(\delta(r) - \theta^k)^2 dr d\theta \tag{10}$$

This implies that the posterior mean $\mathbb{E}[\theta^k|r]$ minimizes the Bayes risk under the squared loss. Thus if the retrieval algorithm could estimate the posterior mean, the theoretical retrieval error will be minimized.

By using Eq. (10) and the universal approximation theorem (UAT) of the ANN (Hornik et al., 1989), we proved that under some mild conditions, the ANN $g(r)$ trained using loss function Eq. (5) converges to the posterior mean estimator $\mathbb{E}[\theta^k|r]$, *i.e.*, $g(r) = \mathbb{E}[\theta^k|r]$. Briefly, we first used large number theory in probability theory to demonstrate that the optimization objective of the ANN Eq. (4) converges to the Bayes risk Eq. (8) with squared loss function, which allows the

ANN to adequately approximate the posterior mean estimator by Eq. (10). Then, we utilize the UAT of ANN to show that the approximation error is sufficiently small. The detailed statement and proof are in appendix A.1.

Combined the Eq.(1), Eq. (2) and $g(r) = \mathbb{E}[\theta^k|r]$, we can rewrite the output of the ANN $g(r)$ as:

$$g(r) = \mathbb{E}[\theta^k|r] = \frac{1}{p(r)} \int_\theta \theta^k p(r|\theta) p(\theta) d\theta \text{ with } p(r|\theta) = \mathcal{N}(f(\theta), \Sigma_r). \tag{11}$$

According to Eq. (11), the training distribution $p(\theta)$ in the training process (Section 2.1.2) acts as the prior distribution and the variance of added Gaussian noise $\Sigma_r$ is the uncertainty of the observation described in Eq. (2).

The proof of Eq. (11) (see appendix A.1) can be applied to any machine learning model with universal approximation capabilities such as ANN or Gaussian process regression with universal kernels (Micchelli et al., 2006). With some relaxation of the error tolerance, this conclusion can be generalized to other machine learning models that do not have the ability of universal approximation. These are collectively referred to as hybrid methods (Verrelst et al., 2015).

**2.2.3 Using ANN to estimate retrieval uncertainty**

In this section, we proposed a novel loss function called variance loss that enables the second ANN $u(r)$ trained with it to output the posterior variance. Let $h(\theta^k) = (\theta^k - \mathbb{E}[\theta^k|r])^2$ in Eq. (9), we could obtain the posterior variance $\mathbb{V}[\theta^k|r]$:

$$\mathbb{V}[\theta^k|r] = \mathbb{E}[(\theta^k - \mathbb{E}[\theta^k|r])^2|r] = \arg\min_\delta \int_\theta p(\theta) \int_r p(r|\theta) l_v(\theta^k, \delta(r)) dr d\theta, \tag{12}$$

where

$$l_v(\theta^k, \delta(r)) = \left((\theta^k - \mathbb{E}[\theta|r])^2 - \delta(r)\right)^2. \tag{13}$$

This means that the posterior variance $\delta(r) = \mathbb{V}[\theta^k|r]$ minimizes the Bayes risk. We call the loss function Eq. (13) the variance loss. Similar to the proof of Eq. (11) in appendix A.1, we could also prove that with the variance loss, the trained ANN $u(r)$ could output the approximation of the posterior variance with input reflectance $r$, i.e., $u(r) = \delta(r) = \mathbb{V}[\theta^k|r]$. The detailed statement and proof could be found in appendix A.2.

Then, we replace the $\mathbb{E}[\theta^k|r]$ in formula Eq. (13) using $g(r)$ using Eq. (11), we get that the ANN $u(r)$ trained with

$$l_u(\theta^k, r) = \left((\theta^k - g(r))^2 - u(r)\right)^2 \tag{14}$$

could also output the approximation of the posterior variance with input reflectance $r$, i.e., $u(r) = \mathbb{V}[\theta^k|r]$. By taking square root of $u(r)$, the standard deviation of posterior distribution could be obtained.

**2.3 The implement of UpNet algorithm**

In this section, we described the implement of UpNet algorithm based on the theory established in Section 2.2. The overall goal of UpNet is (I) to retrieve the BV and (II) to provide corresponding retrieval uncertainty simultaneously. For (I), this could be done typically by training the first ANN $g(r)$ using the squared loss function (5). By substituting loss function (5) into objective function Eq. (3), the objective function of training $g(r)$ is:

$$J_g = \frac{1}{N}\sum_{i=1}^{N}(\theta_i^k - g(r_i))^2. \tag{15}$$

Here, $\theta_i^k$ is the k-th dimension of BVs sample vector $\theta_i$ (the interested BV), $r_i$ is the simulated reflectance that corresponding to $\theta_i$. The Dataset for training $g(r)$ is $D_1 = \{(\theta_1, r_1), (\theta_2, r_2), \cdots, (\theta_N, r_N)\}$, which could be simulated using RTM as described in Section 2.1.2.

For (II), we could train the second ANN $u(r)$ using the loss function (6). With the theory developed in Section 2.2.3, given the reflectance $r$, the trained $u(r)$ would predict the posterior variance. By substituting the loss function (6) into objective function (4), we got that the objective function of the second ANN $u(r)$ is:

$$J_u = \frac{1}{N}\sum_{i=1}^{N}\left(\left(\theta_i^k - g(r_i)\right)^2 - u(r_i)\right)^2. \tag{16}$$

The corresponding dataset for training $u(r)$ is $D_2 = \left\{\left((\theta_1^k - g(r_1)), r_1\right), \left((\theta_2^k - g(r_2)), r_2\right), \cdots, \left((\theta_N^k - g(r_N)), r_N\right)\right\}$. Here, the $\theta_i$ and $r_i$ is the same terms in dataset $D_1$, $g(r_i)$ is the BV retrieved by the first ANN $g(r)$ trained using formula (15) given reflectance $r_i$. Using variance predicted by $u(r)$, by taking the square root, posterior standard deviation could be obtained.

The overall algorithm for training UpNet is shown in Table 2. We firstly train an ANN $g(r)$ to retrieve the BV of interest. Subsequently, we train a second ANN $u(r)$, specifically designed to estimate the retrieval uncertainty. It can be seen that although two ANNs need to be trained in total, since the samples used to train $g(r)$ can still be used to train $u(r)$, there is no need to simulate new data for training $u(r)$. In other words, compared to typical ANN-based method of RTM inversion, UpNet only requires an additional training cost for $u(r)$.

Table 1

Pseudo code for constructing UpNet.

| UpNet algorithm |
|---|
| **For** number of samples **do**:<br>• Sampling the BVs from the prior distribution $\theta_i \sim p(\theta)$.<br>• Sampling the reflectances $r_i = f(\theta_i) + \epsilon_i, \epsilon_i \sim \mathcal{N}(0, \Sigma_r)$.<br>**end for**<br>• Constructing the training dataset $D_1 = \{(\theta_1, r_1), (\theta_2, r_2), \cdots, (\theta_N, r_N)\}$.<br>• Using $\theta_i^k$ denotes the $k$-th dimension of the BV vector $\theta_i$ to represent interested one of the BVs. Training the ANN $g(r)$ using dataset $D_1$ with objective function $J_g = \frac{1}{N}\sum_{i=1}^{N}\left(\theta_i^k - g(r_i)\right)^2$.<br>• Using trained $g(r)$ to predict $\mathbb{E}[\theta^k|r_i]$ for each sampled reflectance $r_i$ in $D_1$ and constructing the dataset $D_2 = \left\{\left((\theta_1^k - g(r_1)), r_1\right), \left((\theta_2^k - g(r_2)), r_2\right), \cdots, \left((\theta_N^k - g(r_N)), r_N\right)\right\}$.<br>• Training the ANN $u(r)$ using dataset $D_2$ with objective function $J_u = \frac{1}{N}\sum_{i=1}^{N}\left(\left(\theta_i^k - g(r_i)\right)^2 - u(r_i)\right)^2$.<br>Then, the ANN $g(r)$ could be used to retrieve the BV and $u(r)$ could be used to estimate the corresponding retrieval uncertainty. |

## III. Experiments

### 3.1 ProSAIL model and simulated dataset

All experiments are based on a vegetation canopy RTM called ProSAIL model, which is a coupling of the leaf optical property model PROSPECT and the canopy reflectance model SAILH (Jacquemoud et al., 2009). Our version couples PROSPECT-5 and SAILH. The inputs to the PROSPECT-5 model are leaf structure parameter (N), chlorophyll a+b content (Cab), dry matter content (Cm), equivalent water thickness (Cw), brown pigments content (Cbrown), carotenoid concentration (Car) and the output are the leaf reflectance and transmittance which are also the input of the SAILH model. Additional inputs of the SAILH model are leaf area index (LAI), average leaf angle (ALA) of an ellipsoidal distribution (leaf inclination distribution function), hot spot parameter (h), solar zenith angle (SZA), viewing zenith angle (VZA), relative azimuth angle (RAA), and the soil reflectance (SR). The SR is provided by a mixture simple model parameterized with a dry-wet soil factor (psoil) and a soil brightness factor (rsoil):

$$r_s = rsoil \times (psoil \times r_w + (1 - psoil) \times r_d) \quad (17)$$

Here, $r_s$ is the soil reflectance, $r_w$ is reflectance of wet soil and $r_d$ is reflectance of dry soil.

**Table 2**

Summary of PROSAIL variables and their sampling distribution or fixed value in training ANN. In the table, $\mathcal{N}(\mu, \sigma^2)$ is used for normal distribution with mean $\mu$ and variance $\sigma^2$, $\mathcal{U}(a,b)$ is used for uniform distribution with lower bound $a$ and upper bound $b$.

| parameters | Description | Unit | Model | Value or distribution |
|---|---|---|---|---|
| N | Leaf structure parameter | [-] | PROSPECT | $\mathcal{U}(1.3, 2.5)$ |
| Cab | Chlorophyll a+b content | ug/cm^2 | PROSPECT | $\mathcal{N}(30, 20^2)$ |
| Car | Carotenoid concentration | ug/cm^2 | PROSPECT | 8 |
| Cw | Equivalent water thickness | g/cm^2 | PROSPECT | $\mathcal{N}(0.02, 0.01^2)$ |
| Cm | Dry matter content | g/cm^2 | PROSPECT | $\mathcal{N}(0.005, 0.001^2)$ |
| h | Hot spot parameter | [-] | SAILH | 0.3 |
| ALA | Average leaf angle | degree | SAILH | $\mathcal{U}(40, 70)$ |
| LAI | Leaf Area Index | [-] | SAILH | $\mathcal{N}(3, 2^2)$ |
| rsoil | Soil brightness factor | [-] | SAILH | 0.8 |
| psoil | Dry-wet soil factor | [-] | SAILH | $\mathcal{U}(0, 1)$ |
| VZA | Viewing zenith angle | degree | SAILH | 0 |
| SZA | Solar zenith angle | degree | SAILH | $\mathcal{U}(40, 70)$ |
| RAA | Viewing zenith angle | degree | SAILH | 0 |

The distributions of BVs for the training set are shown in Table 2, which were derived from a previous global-scale retrieval algorithm research (Estévez et al., 2021; Estévez et al., 2023; Wang et al., 2022). These distributions, often truncated Gaussian or uniform, were adopted to closely mimic real-world distributions. To generate the training dataset, we simulated datasets of Sentinel-2 MSI and Landsat-8 OLI based on their spectral configuration (Table 3) and distributions of BVs. The simulated reflectance was corrupted with multiplicative noise of a level 4% and additive noise with a standard deviation of 0.01 to account for uncertainties introduced by atmospheric correction, signal processing, and other factors (de Sa et al., 2021), formally $r \sim \mathcal{N}(f(\theta), [0.04 f(\theta) + 0.01]^2)$. For the ANN training dataset, we generated 300,000 samples. For the testing dataset, due to the computationally expensive nature of the MCMC method (described in Section 3.3), we generated 300 samples for validation. This dataset is referred to as the simulated dataset.

**Table 3**

**Description of bands configuration of Landsat-8 OLI and Sentinel-2 MSI.**

| Sensor | Bands | Central wavelength (nm) | Bandwidth (nm) |
|---|---|---|---|
| Landsat-8 OLI | Blue | 482 | 60 |
| | Green | 561 | 57 |
| | Red | 655 | 37 |
| | NIR | 865 | 28 |
| | SWIR 1 | 1609 | 85 |
| | SWIR 2 | 2201 | 187 |
| Sentinel-2 MSI | Blue | 490 | 66 |
| | Green | 560 | 36 |
| | Red | 665 | 31 |
| | Red Edge 1 | 705 | 15 |
| | Red Edge 2 | 740 | 15 |
| | Red Edge 3 | 783 | 20 |
| | NIR1 | 842 | 106 |
| | NIR2 | 865 | 21 |
| | SWIR 1 | 1610 | 91 |
| | SWIR 2 | 2190 | 175 |

### 3.2 Real-world dataset

For the ground measurement, we utilized data from multiple measurement campaigns in Europe within the publicly accessible ImagineS dataset (https://fp7-imagines.eu/). These samples were primarily collected from agricultural fields, including wheat, corn, sunflower and so on. The majority of measurements were acquired using LAI-2200 plant canopy analyzers or digital hemispherical photography, providing both effective LAI and true LAI. Since the ProSAIL model does not account for clumping effects (Wang et al., 2022), we employed effective LAI to construct the validation dataset. Most measurement sampling units (ESUs) were designed to be approximately decametric observations, corresponding to Landsat-8 pixels. To match the ground data, we collected Landsat-8 OLI Level-2A surface reflectance products within a five-day window around the ground measurement dates and filtered for high-quality pixels based on the product's quality assurance (QA) band (specifically, we selected pixels with a QA value of 21824, indicating the highest quality). We obtained a total of 164 pairs of ground-measured LAI and Landsat-8 reflectance for validation, with the ANN training dataset being the same as described in Section 3.1. This dataset is referred to as the real-world dataset.

**3.3 Comparation of UpNet and MCMC**

To validate the proposed theory and method, we compared UpNet with the standard Bayesian inference algorithm MCMC. Here, our comparison is divided into two parts. (I) Validating the theoretical basis of UpNet, specifically that the ANN trained with squared loss outputs the posterior mean as formula (11). (II) Validating the accuracy of the retrieval uncertainty (*i.e.*, posterior standard deviation) estimated by UpNet. To achieve this, we use the MCMC algorithm to generate both the posterior mean and posterior uncertainty as relatively accurate reference values, and compare those estimated by UpNet algorithm.

MCMC constructs a Markov chain whose stationary distribution is the posterior distribution, allowing for sampling from the posterior. Based on these samples, the posterior mean and uncertainty can be computed to provide the retrieval result and its uncertainty. For a detailed description of MCMC, please refer to Bishop (2006). In this study, we employed the PyMC4 package (https://www.pymc.io/welcome.html) in python and used the basic Metropolis-Hastings sampler for sampling. For each reflectance, we set the burn-in period of 100 samples followed by 500 samples to obtain posterior samples of BVs. The prior distribution for the inversion was set to be consistent with the training distribution in Table 2. After sampling the posterior samples of BVs using MCMC, we calculated the mean and standard deviation of the samples as the retrieval result and the retrieval uncertainty.

For UpNet, we trained two ANNs following Table 1. They were both designed to have one hidden layer, each with 256 neurons, and a rectified linear unit (ReLU) activation function (Goodfellow et al., 2016). For training, we employed the Adam optimizer with a learning rate of 0.001 (Kingma, 2014), a batch size of 30,000 samples, and 3000 epochs. The regularization coefficient was empirically set to 0.001 to mitigate overfitting and smooth the output results. To ensure numerical stability, we standardized all samples of BVs before both training and predicting of UpNet.

We selected two of the most commonly used BVs as the example: LAI and Cab, which play critical roles in earth science and agriculture (Verrelst et al., 2012; Yan et al., 2019; Lai et al., 2022). For other variables, we did not train the UpNet. For simulated dataset, we trained ANNs to retrieve LAI, Cab and estimate their retrieval uncertainty using both Landsat-8 and Sentinel-2 reflectance. For real-world dataset, due to the fact that the Sentinel-2 launch time is later than the ground measurement time and the absence of field-measured Cab data in dataset, we only utilized Landsat-8 reflectance for retrieving LAI and estimating the retrieval uncertainty. We employed R-squared ($R^2$) and root mean squared error (RMSE) as evaluation metrics to measure the difference between results from UpNet and MCMC. We also compared the speed of retrieving BVs and estimating retrieval uncertainties using both the UpNet and MCMC. To fully leverage the parallel computing capabilities of ANNs, we simultaneously estimated the BVs and the posterior uncertainty for 300,000 pixels, and then calculated the time for one pixel. For MCMC, we directly measured the time taken to sample the posterior distribution of BV and calculate the posterior mean and uncertainty for a single pixel. Since the retrieval time can vary with hardware specifications, as a reference, our experiments were conducted on a machine equipped with an Intel i7-13700K CPU, an NVIDIA GeForce RTX 3090 GPU and 32GB of RAM.

## IV. Results

### 4.1 Comparation of BV retrieved from UpNet method and MCMC on simulated dataset

The BVs retrieved by UpNet and MCMC on simulated datasets of Landsat-8 and Sentinel-2 are correspondingly presented in Fig. 2 and Fig. 3, and a summary of the accuracy is provided in Table 4. The Fig. 2 compare the LAI and Cab estimated by the UpNet with posterior mean of LAI and Cab estimated by MCMC using Landsat-8 reflectance. From Fig. 2 (a) and (c), we observed that UpNet outperformed MCMC in terms of retrieval accuracy for both LAI ($R^2$= 0.74, RMSE = 0.83 for ANN-based method and $R^2$= 0.71, RMSE = 0.87 for MCMC) and Cab ($R^2$= 0.79, RMSE = 8.15 for ANN-based method and $R^2$= 0.77, RMSE = 8.37 for MCMC). Theoretically, MCMC can accurately predict the posterior mean with a sufficiently large number of samples, thereby minimizing errors and being the optimal estimator from Eq. (10). However, due to the high computational complexity of remote sensing inversion using MCMC, only a very limited number of samples can be taken (see Section 3.3), which can lead to larger estimation errors. As a result, ANN's approximation of the posterior mean exhibited higher accuracy in our experiments. From Fig. 2 (b) and (d), the results show no significant bias and are almost distributed along the 1:1 line, indicating a high consistency ($R^2$= 0.97) between the UpNet results and the MCMC results. Since the estimation from MCMC is the posterior mean, thus the results validate the theory that ANN trained with squared loss approximate the posterior mean.

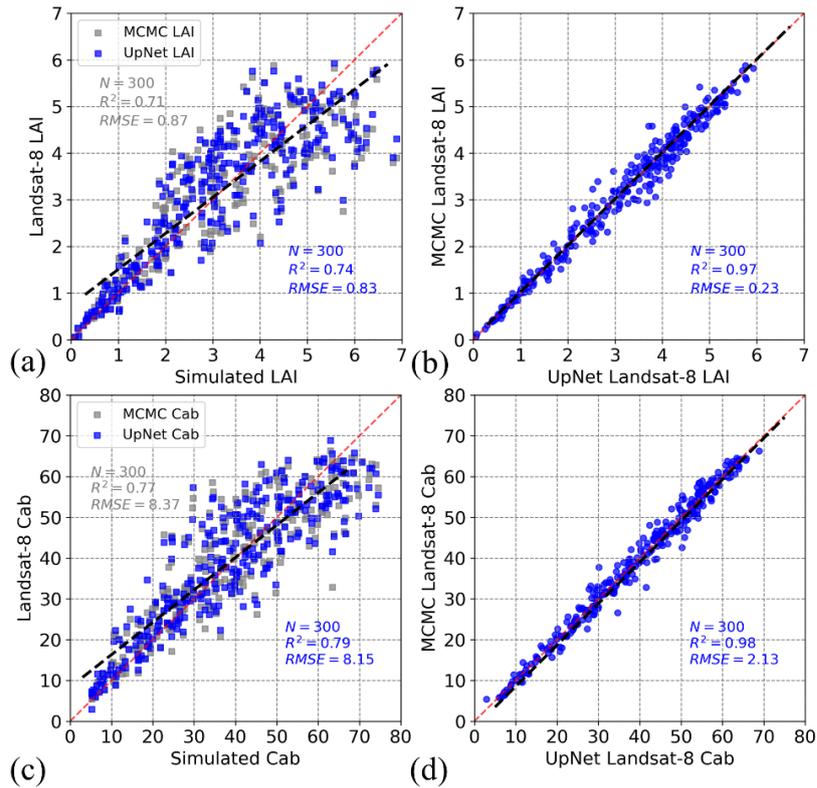

**Fig. 2.** The comparison of BV retrieved by UpNet and MCMC from Landsat-8 reflectance on simulated dataset. The first column presents a comparison of the retrieved (a) LAI and (c) Cab values obtained from both the UpNet and MCMC against the true values, the second column demonstrate the consistency of the retrieved (b) LAI and (d) Cab values obtained from these two methods. The red dashed line is 1:1 line, the black dashed line is the linear regression.

The Fig. 3 compare the LAI and Cab retrieved by the UpNet and MCMC using Sentinel-2 reflectance. Similar with results from Landsat-8, we also find that retrieval accuracy of ANN-based method outperforms MCMC for both LAI ($R^2=$ 0.76, RMSE = 0.79 for ANN-based method and $R^2=$ 0.69, RMSE = 0.92 for MCMC) and Cab ($R^2=$ 0.83, RMSE = 7.3 for ANN-based method and $R^2=$ 0.77, RMSE = 7.93 for MCMC). Moreover, the retrieved BV by UpNet and MCMC is also highly consistent, which further validates the conclusion that the output of ANN trained with squared loss is the posterior mean.

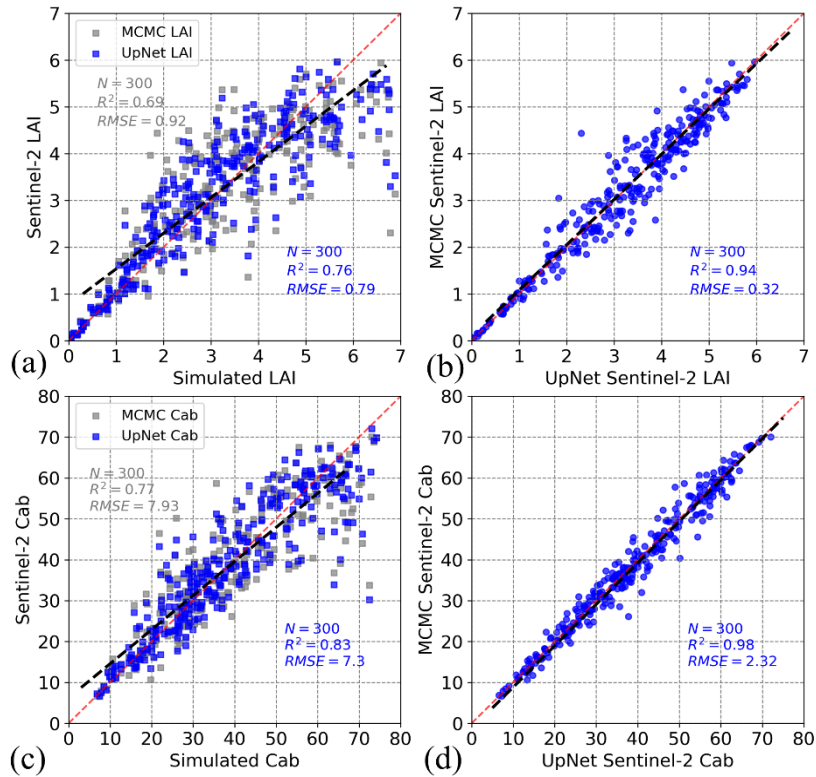

**Fig. 3.** The comparation of BV retrieved by UpNet and MCMC from Sentinel-2 reflectance on simulated dataset. The first column presents a comparison of the retrieved (a) LAI and (c) Cab values obtained from the UpNet and MCMC against the true values, the second column demonstrate the consistency of the retrieved (b) LAI and (d) Cab values obtained from these two methods. The red dashed line is 1:1 line, the black dashed line is the linear regression.

**Table 4** The summary of performance of ANN-based method and MCMC

| Dataset | Sensor | Parameter | Method | $R^2$ | RMSE |
| --- | --- | --- | --- | --- | --- |
| Simulated | Landsat-8 | LAI | UpNet | 0.74 | 0.83 |
| | | | MCMC | 0.71 | 0.87 |
| | | Cab | UpNet | 0.79 | 8.15 |
| | | | MCMC | 0.77 | 8.37 |
| Simulated | Sentinel-2 | LAI | UpNet | 0.76 | 0.79 |
| | | | MCMC | 0.69 | 0.92 |
| | | Cab | UpNet | 0.83 | 7.3 |
| | | | MCMC | 0.77 | 7.93 |
| Real-world | Landsat-8 | LAI | UpNet | 0.64 | 0.81 |
| | | | MCMC | 0.61 | 0.85 |

## 4.2 Comparation of retrieval uncertainty estimated from UpNet and MCMC on simulated dataset

The Fig.4 displays the relationship between the posterior uncertainty estimated by UpNet and that estimated by MCMC. The uncertainties estimated by UpNet are highly correlated with those obtained from MCMC for LAI ($R^2$= 0.84, RMSE = 0.12 for Landsat-8 and $R^2$= 0.56, RMSE = 0.19 for Sentinel-2) and Cab ($R^2$= 0.91, RMSE = 0.99 for Landsat-8 and $R^2$= 0.85, RMSE = 1.08 for MCMC).

We found that when estimating retrieval uncertainty, the consistency between UpNet and MCMC on Sentinel-2 reflectance (Fig. 4 (c) and (d)) is lower compared to that on Landsat-8 reflectance (Fig. 4 (a) and (b)). We attribute this to that Sentinel-2 has more bands, making its MCMC estimation of BVs more difficult. Therefore, under the same sample size, the accuracy will decrease to some extent if the dimension of observations is higher, leading to increased errors in estimating BVs and uncertainties using Sentinel-2 reflectance than using Landsat-8 reflectance.

Moreover, we found that the correlations of uncertainty estimation from UpNet and MCMC (the second column of Fig. 4) are slightly decreased compared to the correlations of BV estimated by these two methods (Fig. 2 (b), (d) and Fig. 3 (b), (d)). These results could be attributed to two factors: (I) the task of estimating uncertainty is more challenging than directly retrieving BVs; (II) the estimation of uncertainty by second ANN in UpNet is based on the posterior mean estimated by the first ANN in UpNet algorithm, meaning that the errors in retrieving BVs will accumulate in the estimation of retrieval uncertainty, leading to a higher error than estimating BV itself.

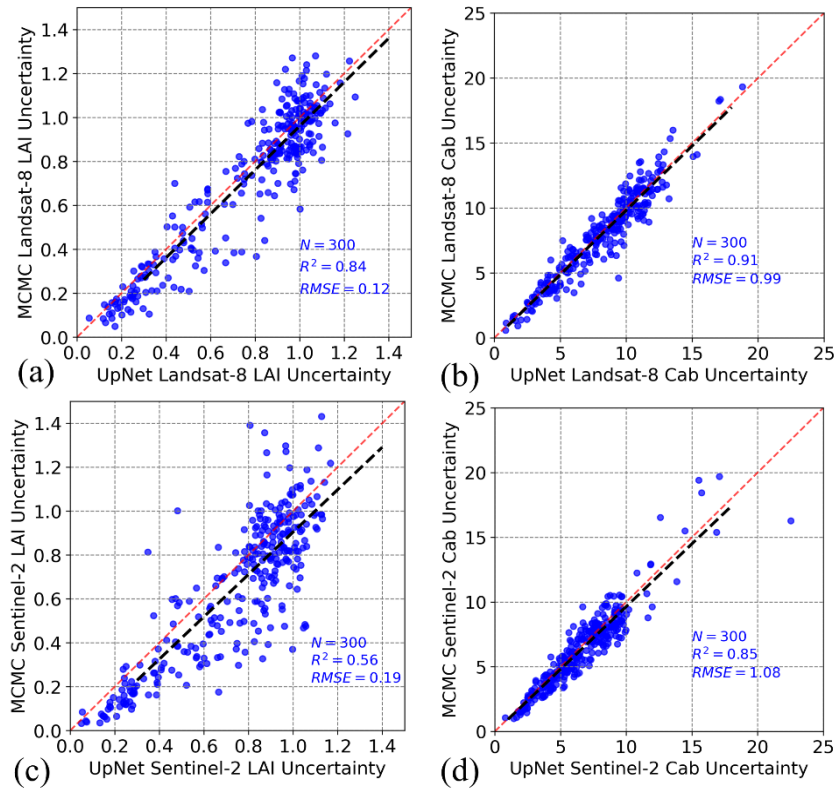

**Fig. 4.** The comparison of retrieval uncertainty estimated by UpNet and MCMC from Landsat-8 and Sentinel-2 reflectance on simulated dataset. Here, (a) and (b) compared the uncertainties estimated by UpNet and MCMC using Landsat-8 reflectance, (c) and (d) compared the uncertainties estimated by UpNet and MCMC using Sentinel-2 reflectance. The red dashed line is 1:1 line, the black dashed line is the linear regression.

## 4.2 Comparation of BV and uncertainty estimated by UpNet and MCMC on real-world dataset

For the real-world dataset, we present both UpNet and MCMC retrieval results along with the retrieval uncertainties in Fig. 5. We found that uncertainty is relatively smaller when LAI values are low, and larger when LAI values are high. When comparing the retrieval accuracy of LAI from UpNet and MCMC estimates, we observed that UpNet slightly outperforms MCMC ($R^2$=0.64 and RMSE = 0.81 with UpNet and $R^2$=0.61 and RMSE = 0.85 with MCMC). The second row of Fig. 5 shows a high consistency between UpNet and MCMC estimates for LAI ($R^2$ = 0.97, RMSE = 0.23) and retrieval uncertainty ($R^2$ = 0.83, RMSE = 0.14). These results provide empirical evidence for the feasibility of UpNet.

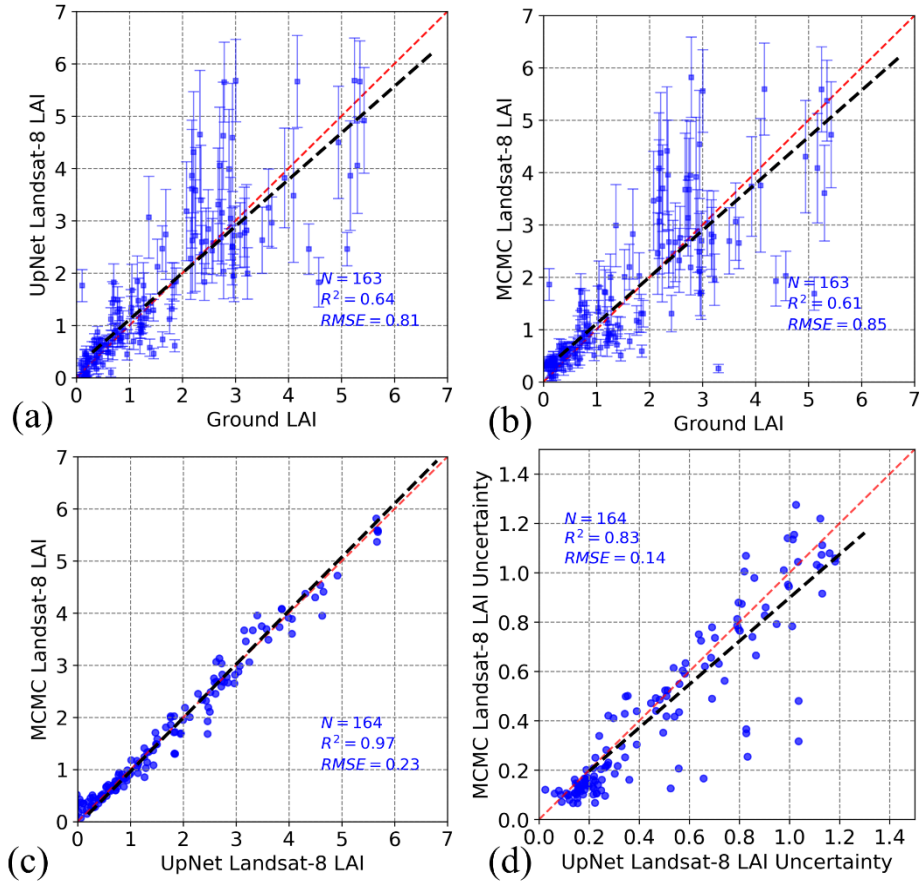

**Fig. 5.** The comparison of retrieval results and retrieval uncertainty using UpNet and MCMC on real-world dataset. Here, (a) and (b) compared the retrieved LAIs from UpNet and MCMC to the ground measured LAI, and (c) and (d) compared the retrieved LAIs and uncertainties from these two methods. The red dashed line is 1:1 line, the black dashed line is the linear regression.

## 4.3 Retrieval Speed of UpNet and MCMC

Table 5 provides a comparison of speed between the UpNet method and MCMC. For the same type of sensor, given the identical ANN architecture, retrieval times are consistent. The results indicate that UpNet achieves speeds over 5,000,000 times faster than MCMC while providing similar or even higher retrieval accuracy. It should be noted that these results are related to the number of samples of posterior sampling for MCMC. If the number of MCMC samples is further increased, MCMC would achieve higher accuracy, but the retrieval time would also increase accordingly.

**Table 5**

**The speed of UpNet and MCMC. "Speed" refers to the total time taken for parameter retrieval and uncertainty estimation.**

| Sensor | Method | Speed for one pixel |
| --- | --- | --- |
| Landsat-8 | UpNet | $6.64 \times 10^{-9}$ second |
|  | MCMC | 38.64 second |
| Sentinel-2 | UpNet | $6.60 \times 10^{-9}$ second |
|  | MCMC | 39.11 second |

## V. Discussion

### 5.1 The characteristics of UpNet

The characteristics of UpNet are simplicity, robustness, and efficiency. Several methods have been proposed to estimate uncertainty using ANNs, including MC dropout (Gal et al., 2016; Martínez-Ferrer et al., 2022, VAE (Kingma, 2013; Svendsen et al., 2024). However, these methods fall short of meeting the requirements of remote sensing applications. MC dropout imposes a Bernoulli distribution prior on model weights and estimates uncertainty by repeatedly sampling weights during retrieval (typically 30-100 times by Gal et al. (2016)). This approach is computationally expensive and has been shown to be unable to output the correct uncertainty of the posterior distribution (Osband, 2016; Folgoc et al., 2021). Variational inference involves complex theoretical underpinnings and relies on a strong Gaussian assumption for the posterior distribution, preventing it from providing the true uncertainty. Furthermore, the training of this method is relatively complex, which hinders its practical application. In contrast, UpNet's theoretical framework does not rely on strong assumptions (*e.g.*, Bernoulli prior on weights or Gaussian assumption of posterior), making it more robust to a wide range of scenarios. And UpNet only requires sequentially training two ANNs, the training method is very straightforward. Another common machine learning model capable of estimating retrieval uncertainty is GPR. However, its computational complexity is $\mathcal{O}(N^3)$ in the number of sample points $N$, significantly limiting its scalability to large datasets (Camps-Valls et al., 2016). UpNet requires only two runs of ANN for BV and uncertainty estimation, its computational complexity is constant as $\mathcal{O}(1)$, which is very computationally efficient, especially when using GPUs. We believe this makes it possible to perform fast BV retrieval and uncertainty estimation on medium and high-resolution images, enabling applications such as data assimilation.

The UpNet also has certain limitations. From an inversion perspective, compared to MCMC, UpNet is less flexible and can only estimate pre-defined statistics such as the mean and variance. Additionally, UpNet suffers from a common problem in ANN-based methods, where it may become unstable when faced with reflectance values not present in the training dataset. In practice, this corresponds to pixels that do not conform to the assumptions of the RTM or have significant noise. To mitigate this issue, we employed a regularization to smooth and constrain the model's output (Section 3.3), but this approach cannot completely address the problem.

### 5.2 The extensibility of UpNet

Using similar method as Section 2.2.3, UpNet can be extended to construct other statistics, such as the covariance between two parameters by modifying the loss functions (As an example, we proof the this in appendix A.3). Furthermore, by training multiple ANNs, we can even predict the covariance matrix of the posterior distribution, allowing for the estimation of correlations in the joint posterior distribution of BVs. In this article, we launch UpNet algorithm with single output ANNs for retrieval and uncertainty quantification for each BV, but UpNet can be generalized to vectors of vegetation parameters with multi-output ANN to accelerate the simultaneous estimation of multiple BVs and providing their retrieval uncertainty.

**5.3 Understand of uncertainty predicted by UpNet**

It is crucial to emphasize that while UpNet algorithm formally utilizes retrieval errors of the first ANN to train the second uncertainty estimation network (see formula (16)), the ANN's objective is to predict posterior variance rather than the error itself. This distinction is essential. The uncertainty represents the distribution of errors in BV retrieval, rather than a specific error value. The error at each point can be considered as a random sample from this error distribution, implying that even for high uncertainty estimates, the error also could be small.

**VI. Conclusion**

In this paper, we proposed a simple, robust, and efficient algorithm called UpNet, which could retrieval BVs and estimating the retrieval uncertainties by training two ANNs. The contributions of this paper can be summarized in three aspects:

(I) Theoretically, to derive UpNet, we first constructed a Bayesian theoretical framework for typical ANN-based methods, demonstrating that ANN trained with squared loss will output the posterior mean of the BV. This serves as the theoretical foundation for UpNet and also provides theoretical guidance for utilizing prior information and regularization in ANN-based methods. Based on this, we introduced a new loss function called variance loss and proved that ANN trained with this loss function can output posterior variance, thereby obtaining posterior uncertainty. These theoretical results provide a solid assurance for the performance of UpNet.

(II) Building on these theories, we constructed the UpNet algorithm to simultaneously retrieve BVs and estimate retrieval uncertainty. Comparisons with the standard Bayesian inference method MCMC demonstrate that our method achieves highly consistent results with MCMC in terms of BV retrieval and the retrieval uncertainty while being over five million times faster. This validates the effectiveness and efficiency of UpNet.

(III) We analyzed the characteristics of UpNet, showing that it has advantages in both theoretical foundation and retrieval speed, as well as good extensibility, allowing for further training to obtain other posterior statistics and multi-dimensional outputs. However, UpNet also has some limitations such as limited flexibility and stability issues.

In summary, our method is supported by a solid theoretical foundation, characterized by simplicity, robustness, and efficiency, enabling fast and accurate retrieval of BVs and provision of retrieval uncertainty in medium and high-resolution remote sensing imagery, with the potential for applying in large-scale BV product production.

**Appendix**

**A.1 Proofs of ANN trained with squared loss predict the posterior mean**

In this appendix, we provide a relatively rigorous mathematical proof for the conclusion that ANN trained with squared loss (5) predict the posterior mean. Firstly, we restate the lemma described in Section 2.2.1.

**Lemma 1** (Lehmann and Casella, 1998). Let the loss function has the form $l_s\left(\theta^k, \delta(r)\right) = \left(\delta(r) - h(\theta^k)\right)^2$, then the estimator $\delta(r) = \mathbb{E}[h(\theta^k)|r]$ minimizes the Bayes risk (8).

Using Lemma 1, we let $h(\theta^k) = \theta^k$ to obtain that $\delta(r) = \mathbb{E}[\theta^k|r]$. Then, we could prove the ANN trained with squared loss outputs the posterior mean. The fundamental idea of our proof is as follows. Firstly, we show that the objective function (4) of the ANN with squared loss is an empirical version of the Bayes risk (8). By the law of large numbers, as the sample size $N$, is large enough, the training objective converges in probability to the Bayes risk, and the weights of the ANN converge to their optimal values, implying that the ANN approximates the optimal ANN which minimize the Bayes risk. Then, we utilize the universal approximation theorem (Hornik et al., 1989) to show that the optimal ANN perfectly approximates the posterior mean estimator. Finally, we employ the triangle inequality to prove that the ANN approximates the posterior mean estimator.

**Theorem 2** (ANN trained with squared loss approximates the posterior mean). We set the ANN $g_\phi$ has continuous activation function, and its weights $\phi$ are in a compact set $\Phi$. Suppose that the optimization algorithm allows the weights $\phi$ of the neural network $g_\phi$ get its optimal $\phi^*$ and training dataset N is large enough, then the ANN $g_\phi$ trained with squared loss converges to the posterior mean estimator in probability. Ignoring the weights $\phi$, we have $g(r) = \mathbb{E}[\theta^k|r]$.

Proof. Using squared loss, formula (8) becomes

$$R(\delta) = \int_\theta p(\theta) \int_r p(r|\theta) (\delta(r) - \theta^k)^2 dr d\theta \qquad (18)$$

The empirical version (Monte Carlo estimation) of Eq. (17) is that:

$$R_{MC}(\delta) = \frac{1}{N}\sum_{i=1}^N \left(\theta_i^k - \delta(r_i)\right)^2.$$

Here, $\theta_i$ is sampled from $p(\theta)$, and $r_i$ is sampled from $p(r|\theta_i)$. We could find that this process is consistent with the training target of the ANN described by formula (4) with squared loss. Using this and viewing the ANN $g_\phi$ as the estimator, it could be show that the $\|R_{MC}(g_\phi) - R(g_\phi)\|_\infty \to 0$ in probability as $N \to \infty$ hold uniformly in $\phi$ (Theorem 9.2 in Keener, (2010)). This implies that when $N$ is large, the optimization objective of the ANN is equivalent to minimizing the Bayes risk. By theorem 9.4 in Keener, (2010) and assumption of optimization algorithm allowing weights of $g_\phi$ find its optimal $\phi^*$, we could get that $\phi \to \phi^*$ in probability. Since $g_\phi$ is continuous, we get that for large N, $|g_\phi - g_{\phi^*}| \le \frac{\epsilon}{2}$ for any small $\epsilon$ in probability. By the universal approximation theorem (Hornik et al., 1989), we can assert that the optimal ANN $g_{\phi^*}$ can approximate the posterior mean estimator $\mathbb{E}[\theta^k|r]$ for any $\epsilon$ with $|g_{\phi^*} - \mathbb{E}[\theta^k|r]| \le \frac{\epsilon}{2}$. Using triangular inequality, we get that for any small $\epsilon$ and large $N$, we have:

$$|g_\phi - \mathbb{E}[\theta^k|r]| \le |g_\phi - g_{\phi^*}| + |g_{\phi^*} - \mathbb{E}[\theta^k|r]| \le \epsilon$$

in probability. Then, by ignoring the weights symbol $\phi$, $g(r) = \mathbb{E}[\theta^k|r]$. This competes the proof.

The Theorem 2 assumed that an ANN can theoretically achieve optimal, though it may not fully realize this in practice.

Nevertheless, attributed to recent advances in optimization, the approximation error of ANN is generally very small. In the proof, we could use the large number theory to state the approximation from $R_{MC}$ to the $R$ due to the samples of BVs $\theta_i$ and reflectances $r_i$ are randomly sampled from the joint distribution $p(\theta, r)$. While equally spaced sampling within the physical range is another common approach, we show that our theoretical framework can accommodate this approach by viewing it as a Riemann integration of $R$ in the parameter space (Zorn, 2002). In some work, the noise is not added, we see this a $\delta$-distribution of the conditional distribution of reflectance $p(r|\theta)$. Since we did not restrict the noise form in our proof, thus these scenarios are fall within the scope of our theoretical framework.

**Corollary 3**. With the same ANN described in Theorem 2, if the ANN $n(r)$ is trained with loss function $l(\theta^k, r) = \left(h(\theta^k) - n(r)\right)^2$, we have $n(r) = \mathbb{E}[h(\theta^k)|r]$.

Proof. The proof is same as the that of Theorem 2, except for replacing the $\theta^k$ to $h(\theta^k)$.

**A.2 Proof of ANN trained with variance loss predict the posterior mean**

In this appendix, we proved that an ANN trained with variance loss predicts the posterior variance.

**Theorem 4.** Assuming the conditions in Theorem 2 hold and using the loss function $l(\theta^k, r) = \left((\theta^k - \mathbb{E}[\theta^k|r])^2 - u(r)\right)^2$, the trained ANN $u(r)$ could output the approximation of the posterior variance given reflectance $r$, formally, $u(r) = \mathbb{E}\left[\theta^k - \mathbb{E}[\theta^k|r]\right]$.

Proof. By using Corollary 3 with $h(\theta) = (\theta^k - \mathbb{E}[\theta^k|r])^2$.

**A.3 Extending the ANN to predict the posterior covariance**

We use $\theta^j$ to represent the $j$-th dimension of BVs vector $\theta$. Then, using Corollary 3 and let $h(\theta) = (\theta^k - \mathbb{E}[\theta^k|r])(\theta^j - \mathbb{E}[\theta^j|r])$, we have $n(r) = \mathbb{E}[(\theta^k - \mathbb{E}[\theta^k|r])(\theta^j - \mathbb{E}[\theta^j|r])|r]$ if the ANN $n(r)$ trained using loss function

$$l(\theta^k, r) = \left((\theta^k - \mathbb{E}[\theta^k|r])(\theta^j - \mathbb{E}[\theta^j|r]) - n(r)\right)^2 \tag{19}$$

**Reference**

Atzberger, C., 2004. Object-based retrieval of biophysical canopy variables using artificial neural nets and radiative transfer models. *Remote sensing of environment*, *93*(1-2), pp.53-67.

Bacour, C., Jacquemoud, S., Leroy, M., Hautecœur, O., Weiss, M., Prévot, L., Bruguier, N. and Chauki, H., 2002. Reliability of the estimation of vegetation characteristics by inversion of three canopy reflectance models on airborne POLDER data. Agronomie, 22(6), pp.555-565.

Baret, F., Hagolle, O., Geiger, B., Bicheron, P., Miras, B., Huc, M., Berthelot, B., Niño, F., Weiss, M., Samain, O. and Roujean, J.L., 2007. LAI, fAPAR and fCover CYCLOPES global products derived from VEGETATION: Part 1: Principles of the algorithm. Remote sensing of environment, 110(3), pp.275-286.

Baret, F. and Buis, S., 2008. Estimating canopy characteristics from remote sensing observations: Review of methods and associated problems. Advances in land remote sensing: System, modeling, inversion and application, pp.173-201.

Bishop, C.M. and Nasrabadi, N.M., 2006. Pattern recognition and machine learning (Vol. 4, No. 4, p. 738). New York: springer.

Camps-Valls, G., Verrelst, J., Munoz-Mari, J., Laparra, V., Mateo-Jimenez, F. and Gomez-Dans, J., 2016. A survey on Gaussian processes for earth-observation data analysis: A comprehensive investigation. IEEE Geoscience and Remote Sensing Magazine, 4(2), pp.58-78.

Combal, B., Baret, F. and Weiss, M., 2002. Improving canopy variables estimation from remote sensing data by exploiting ancillary information. Case study on sugar beet canopies. Agronomie, 22(2), pp.205-215.

Combal, B., Baret, F., Weiss, M., Trubuil, A., Macé, D., Pragnere, A., Myneni, R., Knyazikhin, Y. and Wang, L., 2003. Retrieval of canopy biophysical variables from bidirectional reflectance: Using prior information to solve the ill-posed inverse problem. Remote sensing of environment, 84(1), pp.1-15.

Chen, Y. and Tao, F., 2020. Improving the practicability of remote sensing data-assimilation-based crop yield estimations over a large area using a spatial assimilation algorithm and ensemble assimilation strategies. Agricultural and Forest Meteorology, 291, p.108082.

Danner, M., Berger, K., Wocher, M., Mauser, W. and Hank, T., 2021. Efficient RTM-based training of machine learning regression algorithms to quantify biophysical & biochemical traits of agricultural crops. ISPRS Journal of Photogrammetry and Remote sensing, 173, pp.278-296.

de Sa, N.C., Baratchi, M., Hauser, L.T. and van Bodegom, P., 2021. Exploring the impact of noise on hybrid inversion of PROSAIL RTM on Sentinel-2 data. Remote Sensing, 13(4), p.648.

Disney, M., Muller, J.P., Kharbouche, S., Kaminski, T., Voßbeck, M., Lewis, P. and Pinty, B., 2016. A new global fAPAR and LAI dataset derived from optimal albedo estimates: Comparison with MODIS products. Remote Sensing, 8(4), p.275.

Estévez, J., Berger, K., Vicent, J., Rivera-Caicedo, J.P., Wocher, M. and Verrelst, J., 2021. Top-of-atmosphere retrieval of multiple crop traits using variational heteroscedastic Gaussian processes within a hybrid workflow. Remote sensing, 13(8), p.1589.

Estévez, J., Salinero-Delgado, M., Berger, K., Pipia, L., Rivera-Caicedo, J.P., Wocher, M., Reyes-Muñoz, P., Tagliabue, G., Boschetti, M. and Verrelst, J., 2022. Gaussian processes retrieval of crop traits in Google Earth Engine based on Sentinel-2 top-of-atmosphere data. Remote sensing of environment, 273, p.112958.

Fang, H., Liang, S. and Kuusk, A., 2003. Retrieving leaf area index using a genetic algorithm with a canopy radiative transfer model. Remote sensing of environment, 85(3), pp.257-270.

Fang, H., Wei, S., Jiang, C. and Scipal, K., 2012. Theoretical uncertainty analysis of global MODIS, CYCLOPES, and GLOBCARBON LAI products using a triple collocation method. Remote sensing of environment, 124, pp.610-621.

Fang, H., Jiang, C., Li, W., Wei, S., Baret, F., Chen, J.M., Garcia-Haro, J., Liang, S., Liu, R., Myneni, R.B. and Pinty, B., 2013. Characterization and intercomparison of global moderate resolution leaf area index (LAI) products: Analysis of climatologies and theoretical uncertainties. Journal of Geophysical Research: Biogeosciences, 118(2), pp.529-548.

Fang, H., Baret, F., Plummer, S. and Schaepman-Strub, G., 2019. An overview of global leaf area index (LAI): Methods, products, validation, and applications. Reviews of Geophysics, 57(3), pp.739-799.

Fang, H., Wang, Y., Zhang, Y. and Li, S., 2021. Long-Term variation of global GEOV2 and MODIS leaf area index (LAI) and their uncertainties: An insight into the product stabilities. Journal of Remote Sensing.

Folgoc, L.L., Baltatzis, V., Desai, S., Devaraj, A., Ellis, S., Manzanera, O.E.M., Nair, A., Qiu, H., Schnabel, J. and Glocker, B., 2021. Is MC dropout bayesian? arXiv preprint arXiv:2110.04286.

Gal, Y. and Ghahramani, Z., 2016, June. Dropout as a bayesian approximation: Representing model uncertainty in deep learning. In international conference on machine learning (pp. 1050-1059). PMLR.

Goodfellow, I., 2016. Deep learning. MIT Press.

Hastings, W.K., 1970. Monte Carlo sampling methods using Markov chains and their applications.

Hornik, K., Stinchcombe, M. and White, H., 1989. Multilayer feedforward networks are universal approximators. Neural networks, 2(5), pp.359-366.

Jacquemoud, S., Verhoef, W., Baret, F., Bacour, C., Zarco-Tejada, P.J., Asner, G.P., François, C. and Ustin, S.L., 2009. PROSPECT+ SAIL models: A review of use for vegetation characterization. Remote sensing of environment, 113, pp.S56-S66.

Jiang, J., Weiss, M., Liu, S. and Baret, F., 2022. Effective GAI is best estimated from reflectance observations as compared to GAI and LAI: Demonstration for wheat and maize crops based on 3D radiative transfer simulations. Field Crops Research, 283, p.108538.

Kaipio, J. and Somersalo, E., 2006. Statistical and computational inverse problems (Vol. 160). Springer Science & Business Media.

Keener, R.W., 2010. Theoretical statistics: Topics for a core course. Springer Science & Business Media.

Kingma, D.P., 2013. Auto-encoding variational bayes. arXiv preprint arXiv:1312.6114.

Kingma, D.P., 2014. Adam: A method for stochastic optimization. arXiv preprint arXiv:1412.6980.

Knyazikhin, Y., Martonchik, J.V., Myneni, R.B., Diner, D.J. and Running, S.W., 1998. Synergistic algorithm for estimating vegetation canopy leaf area index and fraction of absorbed photosynthetically active radiation from MODIS and MISR data. Journal of Geophysical Research: Atmospheres, 103(D24), pp.32257-32275.

Koetz, B., Baret, F., Poilvé, H. and Hill, J., 2005. Use of coupled canopy structure dynamic and radiative transfer models to estimate biophysical canopy characteristics. Remote Sensing of Environment, 95(1), pp.115-124.

Lai, Y., Mu, X., Li, W., Zou, J., Bian, Y., Zhou, K., Hu, R., Li, L., Xie, D. and Yan, G., 2022. Correcting for the clumping effect in leaf area index calculations using one-dimensional fractal dimension. Remote Sensing of Environment, 281, p.113259.

Lehmann, E.L. and Casella, G., 2006. Theory of point estimation. Springer Science & Business Media.

Martínez-Ferrer, L., Moreno-Martínez, Á., Campos-Taberner, M., García-Haro, F.J., Muñoz-Marí, J., Running, S.W., Kimball, J., Clinton, N. and Camps-Valls, G., 2022. Quantifying uncertainty in high resolution biophysical variable retrieval with machine learning. Remote Sensing of Environment, 280, p.113199.

Micchelli, C.A., Xu, Y. and Zhang, H., 2006. Universal Kernels. Journal of Machine Learning Research, 7(12).

Osband, I., 2016, December. Risk versus uncertainty in deep learning: Bayes, bootstrap and the dangers of dropout. In NIPS workshop on bayesian deep learning (Vol. 192). MIT Press.

Pinty, B., Andredakis, I., Clerici, M., Kaminski, T., Taberner, M., Verstraete, M.M., Gobron, N., Plummer, S. and Widlowski, J.L., 2011. Exploiting the MODIS albedos with the Two-stream Inversion Package (JRC-TIP): 1. Effective leaf area index, vegetation, and soil properties. Journal of Geophysical Research: Atmospheres, 116(D9).

Richter, K., Hank, T.B., Vuolo, F., Mauser, W. and D'Urso, G., 2012. Optimal exploitation of the Sentinel-2 spectral capabilities for crop leaf area index mapping. Remote Sensing, 4(3), pp.561-582.

Rodgers, C.D., 2000. Inverse methods for atmospheric sounding: theory and practice (Vol. 2). World scientific.

Shiklomanov, A.N., Dietze, M.C., Viskari, T., Townsend, P.A. and Serbin, S.P., 2016. Quantifying the influences of spectral resolution on uncertainty in leaf trait estimates through a Bayesian approach to RTM inversion. Remote Sensing of Environment, 183, pp.226-238.

Svendsen, D.H., Hernandez-Lobato, D., Martino, L., Laparra, V., Moreno-Martinez, A. and Camps-Valls, G., 2023. Inference over radiative transfer models using variational and expectation maximization methods. Machine Learning, 112(3), pp.921-937.

Tang, Y., Marshall, L., Sharma, A., Ajami, H. and Nott, D.J., 2019. Ecohydrologic error models for improved Bayesian inference in remotely sensed catchments. Water Resources Research, 55(6), pp.4533-4549.

Varvia, P., Rautiainen, M. and Seppänen, A., 2017. Modeling uncertainties in estimation of canopy LAI from hyperspectral remote sensing data–A Bayesian approach. Journal of Quantitative Spectroscopy and Radiative Transfer, 191, pp.19-29.

Verrelst, J., Alonso, L., Caicedo, J.P.R., Moreno, J. and Camps-Valls, G., 2012. Gaussian process retrieval of chlorophyll content from imaging spectroscopy data. IEEE Journal of Selected Topics in Applied Earth Observations and Remote Sensing, 6(2), pp.867-874.

Verrelst, J., Rivera, J.P., Moreno, J. and Camps-Valls, G., 2013. Gaussian processes uncertainty estimates in experimental Sentinel-2 LAI and leaf chlorophyll content retrieval. ISPRS journal of photogrammetry and remote sensing, 86, pp.157-167.

Verrelst, J., Rivera, J.P., Veroustraete, F., Muñoz-Marí, J., Clevers, J.G., Camps-Valls, G. and Moreno, J., 2015. Experimental Sentinel-2 LAI estimation using parametric, non-parametric and physical retrieval methods–A comparison. ISPRS Journal of Photogrammetry and Remote Sensing, 108, pp.260-272.

Verrelst, J., Malenovský, Z., Van der Tol, C., Camps-Valls, G., Gastellu-Etchegorry, J.P., Lewis, P., North, P. and Moreno, J., 2019. Quantifying vegetation biophysical variables from imaging spectroscopy data: A review on retrieval methods. Surveys in Geophysics, 40, pp.589-629.

Viskari, T., Hardiman, B., Desai, A.R. and Dietze, M.C., 2015. Model‐data assimilation of multiple phenological observations to constrain and predict leaf area index. Ecological Applications, 25(2), pp.546-558.

Svendsen, D.H., Hernandez-Lobato, D., Martino, L., Laparra, V., Moreno-Martinez, A. and Camps-Valls, G., 2023. Inference over radiative transfer models using variational and expectation maximization methods. Machine Learning, 112(3), pp.921-937.

Wang, J., Lopez-Lozano, R., Weiss, M., Buis, S., Li, W., Liu, S., Baret, F. and Zhang, J., 2022. Crop specific inversion of PROSAIL to retrieve green area index (GAI) from several decametric satellites using a Bayesian framework. Remote Sensing of Environment, 278, p.113085.

Wang, J., Yan, K., Gao, S., Pu, J., Liu, J., Park, T., Bi, J., Maeda, E.E., Heiskanen, J., Knyazikhin, Y. and Myneni, R.B., 2023. Improving the Quality of MODIS LAI Products by Exploiting Spatiotemporal Correlation Information. IEEE Transactions on Geoscience and Remote Sensing, 61, pp.1-19.

Wasserman, L., 2013. All of statistics: A concise course in statistical inference. Springer Science & Business Media.

Weiss, M., Baret, F., Myneni, R., Pragnère, A. and Knyazikhin, Y., 2000. Investigation of a model inversion technique to estimate canopy biophysical variables from spectral and directional reflectance data. Agronomie, 20(1), pp.3-22.

Weiss, M., Jacob, F. and Duveiller, G., 2020. Remote sensing for agricultural applications: A meta-review. Remote sensing of environment, 236, p.111402.

Yan, G., Hu, R., Luo, J., Weiss, M., Jiang, H., Mu, X., Xie, D. and Zhang, W., 2019. Review of indirect optical measurements of leaf area index: Recent advances, challenges, and perspectives. Agricultural and forest meteorology, 265, pp.390-411.

Yan, K., Pu, J., Park, T., Xu, B., Zeng, Y., Yan, G., Weiss, M., Knyazikhin, Y. and Myneni, R.B., 2021. Performance stability of the MODIS and VIIRS LAI algorithms inferred from analysis of long ti6me series of products. Remote Sensing of Environment, 260, p.112438.

Zemp, M., Chao, Q., Han Dolman, A.J., Herold, M., Krug, T., Speich, S., Suda, K., Thorne, P. and Yu, W., 2022. GCOS 2022 implementation plan.

Zérah, Y., Valero, S. and Inglada, J., 2024. Physics-constrained deep learning for biophysical parameter retrieval from Sentinel-2 images: inversion of the PROSAIL model. Remote Sensing of Environment, 312, p.114309.

Zorn, P., 2002. Calculus from graphical, numerical, and symbolic points of view (Vol. 1). McDougal Littel.